\documentclass[10pt,twocolumn,letterpaper]{article}

\usepackage[pagenumbers]{cvpr} 

\usepackage{graphicx}
\usepackage{amsmath}
\usepackage{amssymb}
\usepackage{booktabs}
\usepackage{multirow}
\usepackage{pifont}
\usepackage{color}
\usepackage{threeparttable}
\usepackage{balance}

\usepackage[pagebackref,breaklinks,colorlinks]{hyperref}

\usepackage[capitalize]{cleveref}
\crefname{section}{Sec.}{Secs.}
\Crefname{section}{Section}{Sections}
\Crefname{table}{Table}{Tables}
\crefname{table}{Tab.}{Tabs.}

\def\cvprPaperID{3373} 
\def\confName{CVPR}
\def\confYear{2022}

\begin{document}

\title{Gait Recognition in the Wild with Dense 3D Representations and A Benchmark}

\author{Jinkai Zheng$^1$\thanks{This work was done when Jinkai Zheng was an intern at Explore Academy of JD.com. }\quad
Xinchen Liu$^2 $\thanks{Corresponding author.} \quad
Wu Liu$^{2\dagger}$ \quad
Lingxiao He$^{2}$\quad
Chenggang Yan$^1$ \quad
Tao Mei$^2$\\
$^1$Hangzhou Dianzi University, Hangzhou, China \quad
$^2$Explore Academy of JD.com, Beijing, China \\
{\tt\small \texttt{\{zhengjinkai3, cgyan\}@hdu.edu.cn, \{liuxinchen1, liuwu1, helingxiao3, tmei\}@jd.com}}
}

\maketitle

\begin{abstract}
Existing studies for gait recognition are dominated by 2D representations like the silhouette or skeleton of the human body in constrained scenes. 
However, humans live and walk in the unconstrained 3D space, so projecting the 3D human body onto the 2D plane will discard a lot of crucial information like the viewpoint, shape, and dynamics for gait recognition. 
Therefore, this paper aims to explore dense 3D representations for gait recognition in the wild, which is a practical yet neglected problem.
In particular, we propose a novel framework to explore the 3D Skinned Multi-Person Linear (SMPL) model of the human body for gait recognition, named \textbf{SMPLGait}.
Our framework has two elaborately-designed branches of which one extracts appearance features from silhouettes, the other learns knowledge of 3D viewpoints and shapes from the 3D SMPL model.
In addition, due to the lack of suitable datasets, we build the first large-scale 3D representation-based gait recognition dataset, named \textbf{Gait3D}.
It contains 4,000 subjects and over 25,000 sequences extracted from 39 cameras in an unconstrained indoor scene.
More importantly, it provides 3D SMPL models recovered from video frames which can provide dense 3D information of body shape, viewpoint, and dynamics.
Based on Gait3D, we comprehensively compare our method with existing gait recognition approaches, which reflects the superior performance of our framework and the potential of 3D representations for gait recognition in the wild. 
The code and dataset are available at \href{https://gait3d.github.io}{https://gait3d.github.io}.
\end{abstract}

\section{Introduction}
\label{sec:intro}
\begin{figure}[t]
  \centering
   \includegraphics[width=0.95\linewidth]{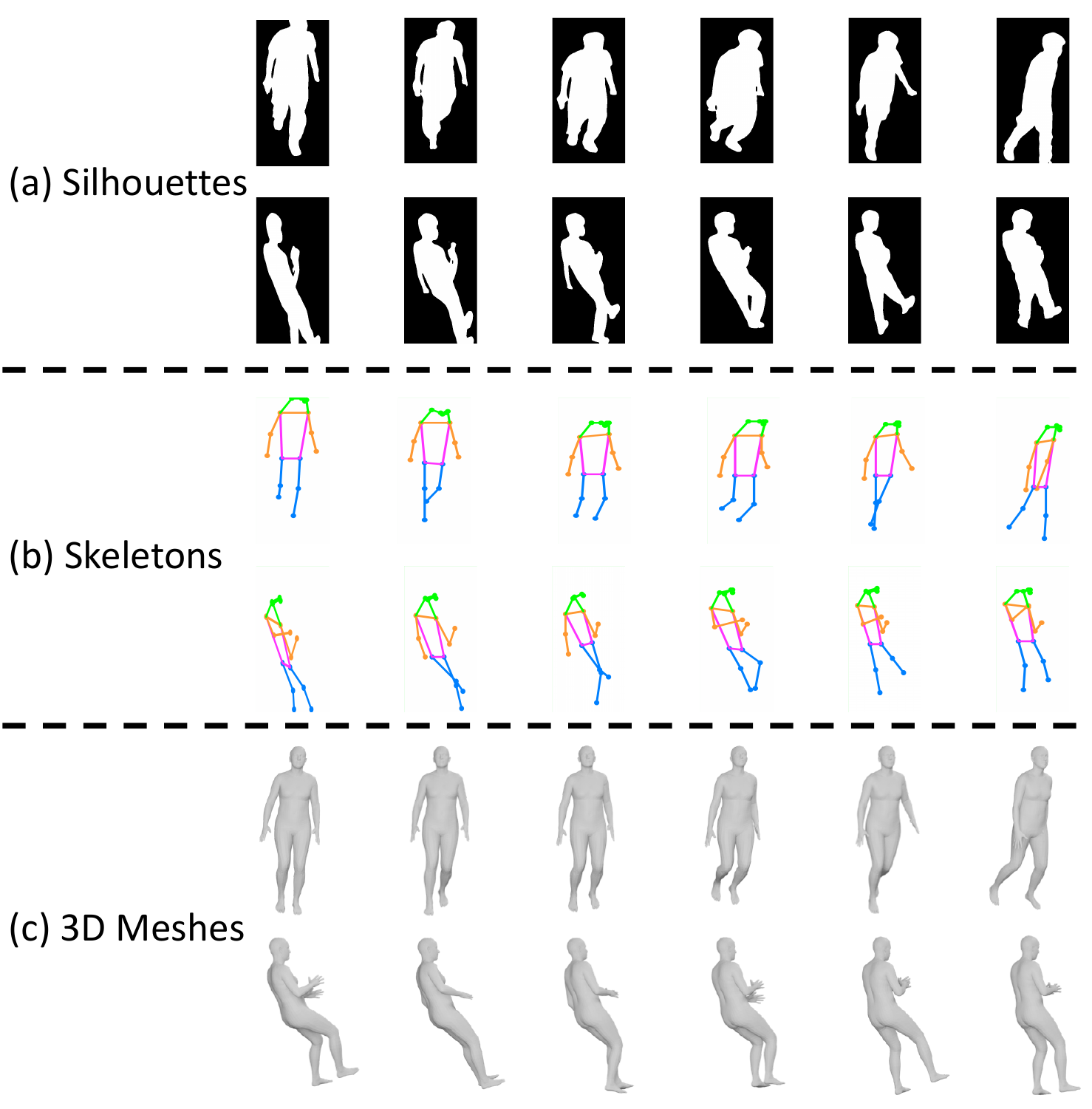}
   \caption{Different gait representations of the same person from two viewpoints. Compared with silhouettes and skeletons, 3D meshes retain the shapes and viewpoints of the human body in the 3D space. (Best viewed in color.)
   }
   \label{fig:figure1} \vspace{-5mm}
\end{figure}

Visual gait recognition, which aims to identify a target person using her/his walking pattern in a video, has been studied for over two decades~\cite{cvpr/NiyogiA94, csur/WanWP19}.
Existing approaches and datasets are dominated by 2D gait representations such as silhouette sequences~\cite{icpr/YuTT06}, Gait Energy Images (GEIs)~\cite{pami/HanB06}, 2D skeletons~\cite{Zhu_2021_ICCV}, as shown in Figure~\ref{fig:figure1}.
However, the human body is a 3D non-rigid object, so the 3D-to-2D projection discards a lot of useful information about shapes, viewpoints, and dynamics while presenting ambiguity for gait recognition.
Therefore, this paper is focused on 3D gait recognition which is valuable yet neglected by the community.

Recently, deep learning-based methods have dominated the state-of-the-art performance on the widely adopted 2D gait recognition benchmarks like CASIA-B~\cite{icpr/TanHYT06} and OU-MVLP~\cite{ipsjtcva/TakemuraMMEY18} by directly learning discriminative features from silhouette sequences~\cite{aaai/ChaoHZF19, cvpr/FanPC0HCHLH20, cvpr/ZhangWL21} or GEIs~\cite{pami/WuHWWT17}.
Despite the excellent results on the in-the-lab datasets, these methods cannot work well in the wild scenarios which have more diverse 3D viewpoints of cameras and more complex environmental interference factors like occlusions~\cite{Zhu_2021_ICCV}.
Although several works exploit 3D cylinders~\cite{icb/AriyantoN11} or 3D skeletons~\cite{fgr/UrtasunF04}, these sparse 3D models also lose helpful information of human bodies like viewpoints and shapes.
Fortunately, the development of parameterized human body models like the Skinned Multi-Person Linear (SMPL) model~\cite{tog/LoperM0PB15} and 3D human mesh recovery approaches~\cite{cvpr/KanazawaBJM18, cvpr/KocabasAB20, Sun_2021_ICCV} makes it possible to estimate precise 3D meshes and viewpoints of human bodies in video frames.
The advantages of 3D meshes for gait recognition are two-fold:
1) the 3D mesh can provide not only the pose but also the shape of the human body in the 3D space, which is crucial for learning discriminative features of gait, and
2) the 3D viewpoint can be explored to normalize the orientations of human bodies during cross-view matching.

To this end, we design a novel 3D SMPL model-based Gait recognition framework, i.e., \textbf{SMPLGait}, to explore the 3D gait representations for human identification.
Our SMPLGait framework has two branches based on deep neural networks.
One branch takes the silhouette sequence of a person as the input to learn appearance features like clothing, hairstyle, and belongings.
However, due to the extreme viewpoint changes in the wild, the shape of the human body can be distorted, which makes the appearance ambiguous, as shown in Figure~\ref{fig:figure1}.
To overcome this challenge, we design a 3D Spatial-Transformation Network (3D-STN) as the other branch to learn 3D knowledge of viewpoint and shape from the 3D human mesh.
The 3D-STN takes the 3D SMPL model of each frame as the input to learn a spatial transformation matrix.
By applying the spatial transformation matrix to the appearance features, these features from different viewpoints are normalized in the latent space.
By this means, the gait sequences of the same person will be closer in the feature space.

Nevertheless, there is no suitable dataset that provides 3D meshes of human bodies in the wild.
Therefore, to facilitate the research, we build the first large-scale 3D mesh-based gait recognition dataset, named \textbf{Gait3D}, from high-resolution videos captured in the wild.
Compared to existing datasets listed in Table~\ref{tab:table1}, the Gait3D dataset has the following featured properties:
\textbf{1)} Gait3D contains 4,000 subjects with over 25,000 sequences captured by 39 cameras in an unconstrained indoor scene which makes it scalable for research and applications.
\textbf{2)} It provides precise 3D human meshes recovered from video frames which can provide 3D pose and shape of human bodies as well as accurate viewpoint parameters.
\textbf{3)} It also provides conventional 2D silhouettes and keypoints which can be explored for gait recognition with multi-modal data.

In summary, the contributions of this paper are as follows:
\begin{itemize}
  \item We make one of the first attempts toward 3D gait recognition in the real-world scenario, which aims to explore dense 3D representations of the human body for gait recognition.
  \item We propose a novel 3D gait recognition framework based on the SMPL model, named SMPLGait, to explore 3D human meshes for gait recognition.
  \item We build the first large-scale 3D gait recognition dataset, named Gait3D, which provides the 3D human meshes of gait collected from unconstrained scenarios.
\end{itemize}
Through comprehensive experiments, we not only evaluate existing 2D silhouettes/skeleton-based approaches but also demonstrate the effectiveness of the proposed SMPLGait method, which reflects the potential of 3D representations for gait recognition.
Moreover, the combination of 3D and 2D representations further improves the performance which shows the complementarity of multi-modal representations.

\begin{figure*}[t]
  \centering
   \includegraphics[width=0.98\linewidth]{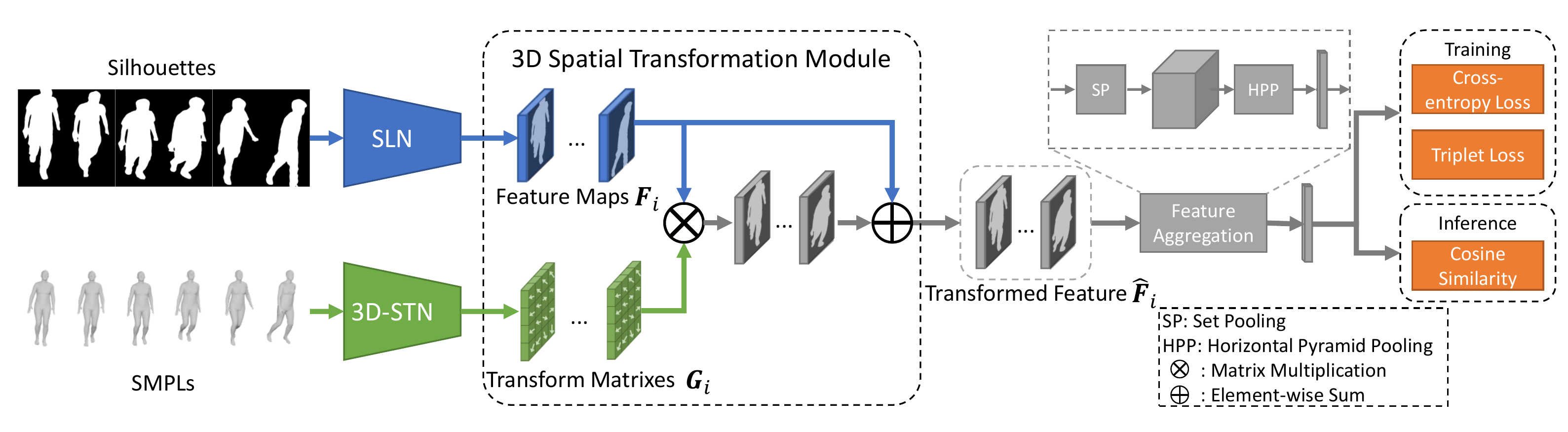}
   \caption{The architecture of the SMPLGait framework for 3D gait recognition in the wild. }
   \label{fig:figure4} \vspace{-5mm}
\end{figure*}

\section{Related Work}
\label{sec:related}
\textbf{Gait Recognition.}
We review the 2D and 3D representations-based gait recognition methods separately.

2D gait recognition methods can be classified into model-based and model-free approaches~\cite{csur/WanWP19}.
Early methods mainly belong to the model-based which defines a structural human body model.
Then, gait patterns are modeled by parameters like lengths of limbs, angles of joints, and relative positions of body parts~\cite{pr/YamNC04, icb/AriyantoN11}.
The model-free methods mainly adopt the silhouettes obtained by background subtraction from video frames~\cite{pami/HanB06, icb/ShiragaMMEY16, pami/WuHWWT17, aaai/ChaoHZF19, cvpr/ZhangT0A0WW19, cvpr/FanPC0HCHLH20, eccv/HouCLH20, cvpr/ZhangWL21, Lin_2021_ICCV, Huang_2021_ICCV}.
In particular, Han~\textit{et al.} proposed to aggregate a sequence of silhouettes into a compact Gait Energy Image (GEI)~\cite{pami/HanB06} which was widely used by the following methods~\cite{icb/ShiragaMMEY16, pami/WuHWWT17}.
Recently, due to the success of deep learning for computer vision tasks~\cite{pami/gongbao_ycg, csvt/ycg_2022, tomm/tengtong_ycg, tomm/lizhisheng_ycg, tomm/ycg_age_2022, liu2021recent, tmm/Liu18provid, icme/Liu16veri, iccv/WuLYLLL19, iccv/ChenLL00021}, deep Convolutional Neural Networks (CNNs) also dominated the performance of gait recognition.
For example, Shiraga~\textit{et al.}~\cite{icb/ShiragaMMEY16} and Wu~\textit{et al.}~\cite{pami/WuHWWT17} proposed to learn effective features from GEIs and significantly outperformed previous methods.
The most recent methods started to learn discriminative features directly from the silhouette sequences using larger CNNs or multi-scale structures and achieved state-of-the-art results~\cite{aaai/ChaoHZF19, cvpr/FanPC0HCHLH20, Huang_2021_ICCV, Lin_2021_ICCV}.
Despite the excellent performance on in-the-lab datasets, e.g., CASIA-B and OU-LP, these methods usually fail in the wild as shown in the experiments on GREW~\cite{Zhu_2021_ICCV} and our Gait3D.

3D representations have also been studied since the early years of gait recognition.
For example, Urtasun and Fua~\cite{fgr/UrtasunF04} proposed an approach to gait analysis that depended on 3D temporal motion models using an articulated skeleton.
Zhao~\textit{et al.}~\cite{fgr/ZhaoLLP06} applied a local optimization algorithm to track 3D motion for gait recognition.
Yamauchi~\textit{et al.}~\cite{cvpr/YamauchiBS09} proposed the first method using 3D pose estimated from RGB frames for walking human recognition.
Ariyanto and Nixon~\cite{icb/AriyantoN11} built a 3D voxel-based dataset using a complex multi-camera system and proposed a structural model of articulated cylinders with 3D Degrees of Freedom at each joint to model the human lower legs.
However, these methods either discard rich 3D information like viewpoints and shapes or are limited by devices for real-world applications.
In summary, to overcome the problem of 2D methods and explore 3D representations for gait recognition in the wild, we aim to explore 3D mesh as a rich representation of the viewpoint and shape of the human body.

\textbf{Gait Recognition Datasets.}
Current publicly available gait recognition datasets mainly belong to two series, i.e., the CASIA series~\cite{pami/WangTNH03, icpr/YuTT06, icpr/TanHYT06} and the OU-ISIR series~\cite{cvpr/TsujiMY10, pr/HossainMWY10, accv/MakiharaMY10, tifs/IwamaOMY12, ipsjtcva/XuMOLYL17, ipsjtcva/UddinTMTLMY18, tbbis/AnYMWXYLY20} as listed in Table~\ref{tab:table1}.
The CASIA series were built in the early research of gait recognition, which facilitated the initial exploration of RGB images and silhouettes for gait representations~\cite{icpr/TanHYT06, pami/HanB06}.
Despite its smaller number of subjects, the CASIA-B~\cite{icpr/YuTT06} is still the most widely used dataset for the evaluation of silhouette-based methods.
The OU-ISIR series were first built ten years ago and developed comprehensive variants such as walking at different speeds~\cite{cvpr/TsujiMY10}, clothing styles~\cite{pr/HossainMWY10}, and bags~\cite{ipsjtcva/UddinTMTLMY18}, subjects of different ages~\cite{ipsjtcva/XuMOLYL17}, and annotations of 2D pose~\cite{tbbis/AnYMWXYLY20}.
Due to their large population, the OU-LP~\cite{tifs/IwamaOMY12} and OU-MVLP~\cite{ipsjtcva/TakemuraMMEY18} also became the most popular datasets for current research.
However, the above datasets were collected in constrained scenes like labs~\cite{icpr/YuTT06, tifs/IwamaOMY12} or a small defined area on a campus~\cite{pami/WangTNH03, pami/SarkarPLVGB05}.
Most recently, researchers started to narrow the gap between in-the-lab research and real-world application.
As a contemporaneous study of our work, Zhu~\textit{et al.}~\cite{Zhu_2021_ICCV} constructed the GREW dataset from natural videos collected in an open area.
However, there is no dataset that provides rich 3D representations for gait recognition in the wild.
Therefore, we need to build a new dataset that is collected from complex scenes and with dense 3D meshes for gait recognition in the wild.

\textbf{3D Human Mesh Recovery.}
3D representations have attracted a lot of attention in the computer vision community~\cite{eccv/ZhouKGLH18ganchuang3D, corr/abs-2203-06558-ganchuang3D}.
The 3D human body can be represented by point clouds~\cite{pami/LiuSPWDK20}, voxels~\cite{fgr/UrtasunF04}, parameterized blend shapes ~\cite{tog/AnguelovSKTRD05}, etc.
Among them, the Skinned Multi-Person Linear (SMPL) model~\cite{tog/LoperM0PB15}
is a skinned vertex-based model that can accurately represent a wide variety of body shapes in natural human poses.
With the SMPL model, an arbitrary 3D human body can be represented by a linear combination of a group of shape, pose, scale, and viewpoint parameters.
Based on the SMPL model, a series of 3D human mesh recovery approaches are developed to estimate accurate 3D shapes, poses, and viewpoints of human bodies from natural images~\cite{cvpr/KanazawaBJM18, cvpr/KocabasAB20, Sun_2021_ICCV, Sun_2022_CVPR_BEV}.
These methods provide us an opportunity to obtain 3D human meshes from in-the-wild videos for 3D mesh-based gait recognition.

\section{The 3D Gait Recognition Method}
\subsection{Overview}
\label{subsec:overview}
The overall architecture of the proposed 3D SMPL-based Gait Recognition framework, SMPLGait, is shown in Figure~\ref{fig:figure4}.
There are two branches of the framework.
For the first branch, we take the sequence of silhouettes as input which has a rich knowledge of the appearance and use a CNN-based model to extract 2D spatial features from each frame.
For the second branch, the SMPLs of the human body are fed into the 3D-Spatial-Transformation Network (3D-STN), which aims to learn the latent transformation matrixes from the 3D viewpoints and shapes.
Then the 3D Spatial Transformation Module aligns the 2D appearance features in the latent space using the learned transformation matrixes.
Finally, the transformed feature of each frame is aggregated into a sequence-level feature for sequence-to-sequence matching in training or inference.
Next, we will introduce the above modules in detail.

\begin{table*}[t]
\begin{center}
\begin{threeparttable}
\small
\begin{tabular}{lcccccccc}
Dataset        						        & Year  & Subject \#    & Seq \# & Cam \# & Data Type     & Speed & Wild & 3D-View\\  
\midrule[1.5pt]
CASIA-A~\cite{pami/WangTNH03} 	            & 2003	& 20	        & 240       & 3	    & RGB, Silh.	   & \textcolor{red}{\ding{55}} & \textcolor{red}{\ding{55}} & \textcolor{red}{\ding{55}}\\
USF HumanID~\cite{pami/SarkarPLVGB05} 	    & 2005  & 122		    & 1,870     & 2	    & RGB              & \textcolor{red}{\ding{55}} & \textcolor{red}{\ding{55}} & \textcolor{red}{\ding{55}}\\
CASIA-B~\cite{icpr/YuTT06} 	                & 2006	& 124		    & 13,640    & 11	& RGB, Silh.       & \textcolor{red}{\ding{55}} & \textcolor{red}{\ding{55}} & \textcolor{red}{\ding{55}}\\
CASIA-C~\cite{icpr/TanHYT06} 	            & 2006	& 153		    & 1,530     & 1	    & Infrared, Silh.  & \textcolor{green}{\ding{51}} & \textcolor{red}{\ding{55}} & \textcolor{red}{\ding{55}}\\
OU-ISIR Speed~\cite{cvpr/TsujiMY10} 	    & 2010 	& 34		    & 306       & 1	    & Silh.            & \textcolor{green}{\ding{51}} & \textcolor{red}{\ding{55}} & \textcolor{red}{\ding{55}}\\
OU-ISIR-LP~\cite{tifs/IwamaOMY12}		        & 2012	& 4007          & 31,368    & 2	    & Silh.            & \textcolor{red}{\ding{55}} & \textcolor{red}{\ding{55}} & \textcolor{red}{\ding{55}}\\ 
OU-LP Bag~\cite{ipsjtcva/UddinTMTLMY18}     & 2018	& 62,528        & 187,584   & 1	    & Silh.            & \textcolor{red}{\ding{55}} & \textcolor{red}{\ding{55}} & \textcolor{red}{\ding{55}}\\
OU-MVLP~\cite{ipsjtcva/TakemuraMMEY18}	    & 2018	& 10,307        & 288,596   & 14	& Silh.            & \textcolor{red}{\ding{55}} & \textcolor{red}{\ding{55}} & \textcolor{red}{\ding{55}}\\ 
OU-MVLP Pose~\cite{tbbis/AnYMWXYLY20}	    & 2020	& 10,307        & 288,596   & 14    & 2D Pose          & \textcolor{red}{\ding{55}} & \textcolor{red}{\ding{55}} & \textcolor{red}{\ding{55}}\\ 
GREW~\cite{Zhu_2021_ICCV}	                & 2021	& 26,345        & 128,671   & 882   & Silh., 2D/3D Pose, Flow & \textcolor{red}{\ding{55}} & \textcolor{green}{\ding{51}} & \textcolor{red}{\ding{55}}\\ 
\midrule
\textbf{Gait3D}							    & - 	& \textbf{4,000}         & \textbf{25,309}    & \textbf{39}    & \textbf{Silh., 2D/3D Pose, 3D Mesh\&SMPL} & \textcolor{green}{\ding{51}} & \textcolor{green}{\ding{51}} & \textcolor{green}{\ding{51}}\\
\end{tabular}
\end{threeparttable} \vspace{-3mm}
\caption{Comparison of publicly available datasets for gait recognition. Speed, Wild, and 3D-View indicate whether the dataset contains inconstant walking speed, is captured in the wild, and has viewpoint variations in the 3D space, respectively.}
\label{tab:table1}
\end{center} \vspace{-9mm}
\end{table*}

\subsection{Network Structure}
\label{subsec:network}
\textbf{The Silhouette Learning Network} (SLN) aims to learn the appearance knowledge of humans from silhouettes that contain 2D spatial information like clothing and hairstyle.
The SLN has six convolutional layers which are similar to the backbone of GaitSet~\cite{aaai/ChaoHZF19}.
As is shown in Figure~\ref{fig:figure4}, the sequences of silhouettes are fed into a CNN.
We formulate $X_{sil} = \{ \mathbf{x}_i\}_{i=1}^L$ as the input sequence, where $ \mathbf{x}_i \in \mathbb{R}^{H \times W}$ is the $i$-th binary frame, $L$ is the length of the sequence, H and W are the height and width of the silhouette image.
For a frame $\mathbf{x}_i$, the process can be formulated as:
\begin{equation}
\label{equ1}
\mathbf{F}_i = F(\mathbf{x}_i),
\end{equation}
where $F(\cdot)$ is the CNN-based backbone and $\mathbf{F}_i \in \mathbb{R}^{h \times w}$ is the frame-level feature map for frame $\mathbf{x}_i$~\footnote{For the convenience of notation, we omit the channel of the feature map.}.

\textbf{The 3D Spatial Transformation Network} (3D-STN) is proposed to solve viewpoint changes in real 3D scenarios.
3D SMPL parameters related to 3D viewpoints, shapes, and poses are the input of this module.
Assuming $Y_{sp} = \{ \mathbf{y}_i \}_{i=1}^L$ is the input SMPLs, where $ \mathbf{y}_i \in \mathbb{R}^D $ is the SMPL vector of $i$-th frame, $D$ is the dimension of the SMPL vector which contains 24$\times$3 dimensions of 3D human body pose, 10 dimensions of 3D body shape, and 3 dimensions of camera scale and translation parameters.
The 3D-STN consists of three fully connected (FC) layers with neuron number $=128 \Rightarrow 256 \Rightarrow h \times w$, where $h$ and $w$ are the height and width of the feature map from the Silhouette Learning Network.
Each FC layer is followed by batch normalization and the ReLU activation function.
We use dropout for the last two FC layers to eliminate overfitting.
The forward process of 3D-STN can be formulated as:
\begin{equation}
\label{equ2}
\mathbf{g}_i = G(\mathbf{y}_i),
\end{equation}
where $G(\cdot)$ is the 3D-STN and $\mathbf{g}_i$ is the frame-level transformation vector for frame $i$.

\textbf{The 3D Spatial Transformation Module} is designed to align the 2D appearance feature map $\mathbf{F}_i \in \mathbb{R}^{h \times w}$ using the transformation vector $\mathbf{g}_i$ in the feature space, as shown in Figure~\ref{fig:figure4}.
We first reshape the transformation vector $\mathbf{g}_i$ to a matrix $\mathbf{G}_i \in \mathbb{R}^{w \times h}$.
Then, for convenience of computation, we expand $\mathbf{F}_i$ and $\mathbf{G}_i$ to square matrixes by zero padding on the short edge.
After that, we apply $\mathbf{G}_i$ to $\mathbf{F}_i$ by
\begin{equation}
\label{equ3}
\widehat{\mathbf{F}}_i = \mathbf{F}_i \cdot (I + \mathbf{G}_i),
\end{equation}
where $I$ is an identity matrix and $\cdot$ is matrix multiplication.
At last, we adopt Set Pooling (SP) and Horizontal Pyramid Pooling (HPP) in GaitSet~\cite{aaai/ChaoHZF19} to aggregate $\widehat{\mathbf{F}_i}$ into the final feature vector for sequence-to-sequence matching.
For more details of the SMPLGait framework, please refer to \textbf{the supplementary material}.

\subsection{Training and Inference}
\label{subsec:train}
Our two-branch 3D gait recognition framework is trained in an end-to-end manner.
The network of our framework is optimized by a loss function with two components:
\begin{equation}
\label{equ4}
L=\alpha L_{tri}+\beta L_{ce},
\end{equation}
where $L_{tri}$ is the triplet loss, $L_{ce}$ is the cross entropy loss.
$\alpha$ and $\beta$ are the weighting parameters. 

During inference, we use the sequences of silhouettes and SMPLs as the inputs of the two branches, respectively.
The cosine similarity is used to measure the similarity between a query-gallery pair.

\section{The Gait3D Benchmark}
\label{sec:dataset}

To facilitate the research of 3D gait recognition, we present a novel large-scale dataset, named Gait3D, which has several featured properties compared to existing datasets in Table~\ref{tab:table1}.
First of all, the Gait3D dataset consists of 4,000 subjects, 25,000+ sequences, and over 3 million bounding boxes captured by cameras of arbitrary 3D viewpoints, which makes it more scalable for training deep CNNs.
Moreover, it provides accurate 3D human meshes estimated from video frames, which contains the poses and shapes of human bodies as well as viewpoints in the 3D space.
Furthermore, Gait3D also provides 2D silhouettes and 2D/3D keypoints obtained by the state-of-the-art image segmentation and pose estimation methods fine-tuned on our dataset.
Therefore, multi-modal data can be explored for gait recognition.
In addition, Gait3D is collected in a large supermarket in which people usually walk at irregular speeds and routes, and can be occluded by other people or objects.
The above properties also make Gait3D a scalable but challenging dataset for gait recognition which can be reflected by the evaluation in Section~\ref{sec:experiments}.

\subsection{Data Collection and Pre-processing}
\label{subsec:datacollect}
To collect a high-quality in-the-wild dataset for real applications, we collect the seven-day raw videos from 39 cameras mounted in a large supermarket.
The scenes of the cameras include the entrance, the goods shelf area, the freezer area, the dining area, the checkout counter, etc.
For the videos each day, we randomly sample two segments of continuous two-hour videos.
At last, we obtain about 1,090 hours videos with 1,920 $\times$ 1,080 resolution and 25 FPS.
Note that, we are authorized by the management of the supermarket to access and process the data for research purposes.
In addition, all subjects were noticed that the data is collected only for research purposes.
With the videos, we use the open-source FFmpeg~\footnote{\url{http://ffmpeg.org/} under the GNU LGPL License v2.1.} to decode the raw videos into frames at 25 FPS to keep the continuity of gait sequences.
To guarantee the high quality of the dataset, the annotation process is performed in three main steps as follows.

\subsection{Dataset Construction}
\label{subsec:construct}

\subsubsection{Person detection and tracking from frames}
For each frame extracted from the raw videos, we adopt the CenterNet~\cite{corr/abs-1904-07850} fine-tuned on our dataset as the person detector since it is an efficient anchor-free object detector~\footnote{4,000 person bounding boxes are labeled for fine-tuning the detector.}.
To achieve accurate person tracking in videos, we exploit the Intersection-over-Union (IoU) and person re-identification (ReID) features of bounding boxes in two adjacent frames to measure their similarity.
The ReID feature is extracted by an open-source person ReID framework, FastReID~\footnote{\url{https://github.com/JDAI-CV/fast-reid} under the Apache 2.0 license.}~\cite{corr/abs-2006-02631} pretrained on several public person ReID datasets.
When two persons are highly overlapped, the tracking algorithm can easily misjudge them as one person, i.e., ID switching. 
To solve this problem, we employ human annotators to clean sequences that may contain more than one pedestrian. 
By this means, we guarantee that each sequence only belongs to one person.
Then, we discard the sequences shorter than 25 frames or longer than 500 frames and obtain about 50,000 sequences in total.

\subsubsection{Cross-camera sequence matching}
With the above sequences, we should cluster the sequences of the same person in all cameras.
To achieve effective and efficient cross-camera matching of the same person, we also utilize the person ReID features obtained by FastReID~\cite{corr/abs-2006-02631}.
For each sequence, we first use a pose estimation model, i.e., HRNet~\footnote{\url{https://github.com/HRNet/HRNet-Human-Pose-Estimation} under the MIT License.}~\cite{pami/00010CJDZ0MTW0X21} fine-tuned on our dataset~\footnote{4,000 images are labeled to fine-tune the pose estimator.}, to select a high-quality frame for cross-camera matching.
After that, we utilize FastReID to extract the features of the selected frames of all sequences.
Through an unsupervised clustering method, i.e., DBSCAN~\cite{kdd/EsterKSX96}, we roughly obtain 5,336 clusters of sequences.
Then, we employ human annotators to filter out the outlier sequences in each group.
By discarding the groups containing only one sequence, we finally obtain 4,000 subjects and 25,309 sequences for generating the gait representations.

\begin{figure}[t]
  \centering
   \includegraphics[width=0.95\linewidth]{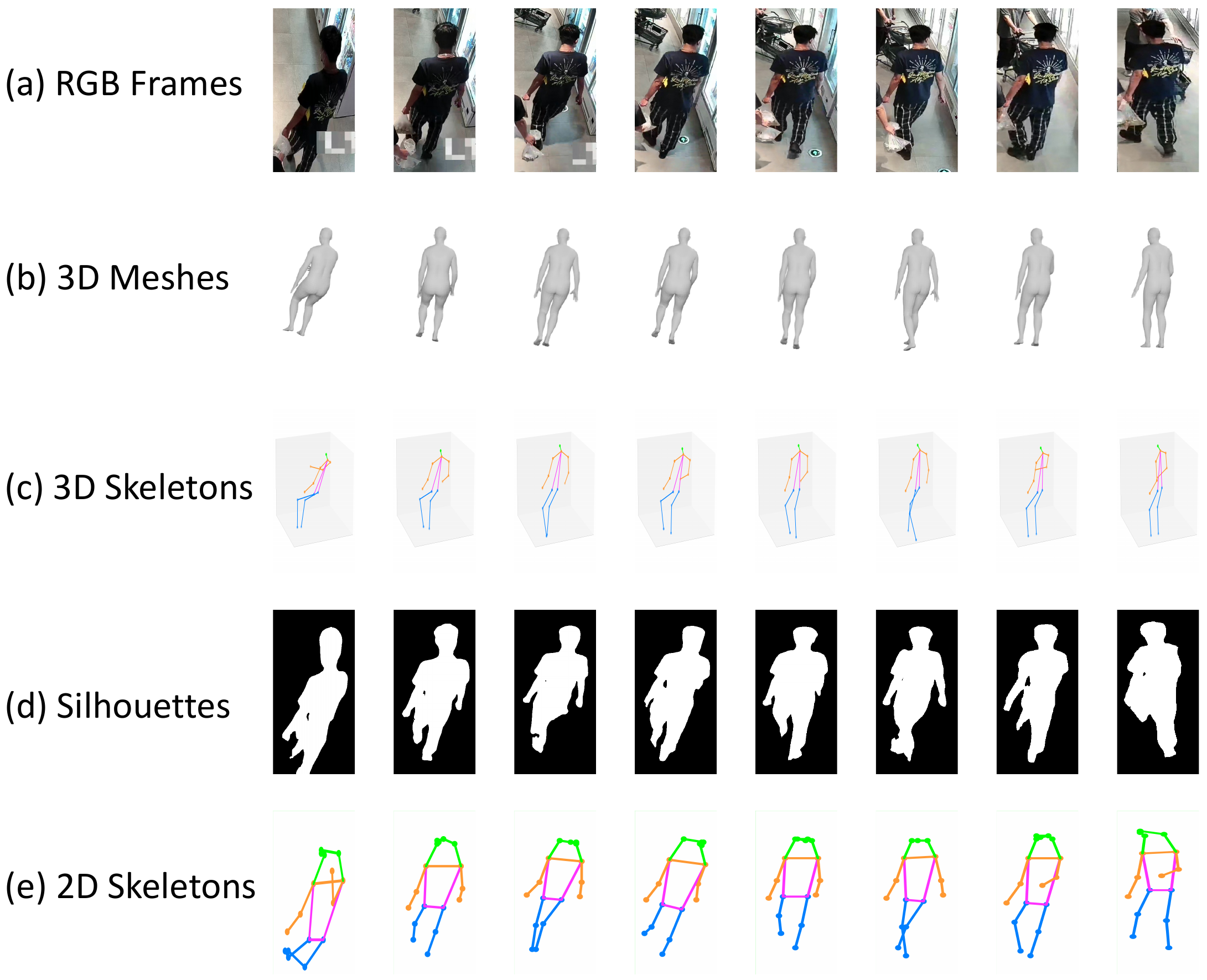}
   \caption{Examples of gait representations in the Gait3D dataset. The sizes are normalized for visualization. (Best viewed in color.)}
   \label{fig:figure2} \vspace{-5mm}
\end{figure}

\begin{figure*}[t]
  \centering
  \begin{minipage}[t]{0.3\linewidth}
  \centering
  \includegraphics[width=\linewidth]{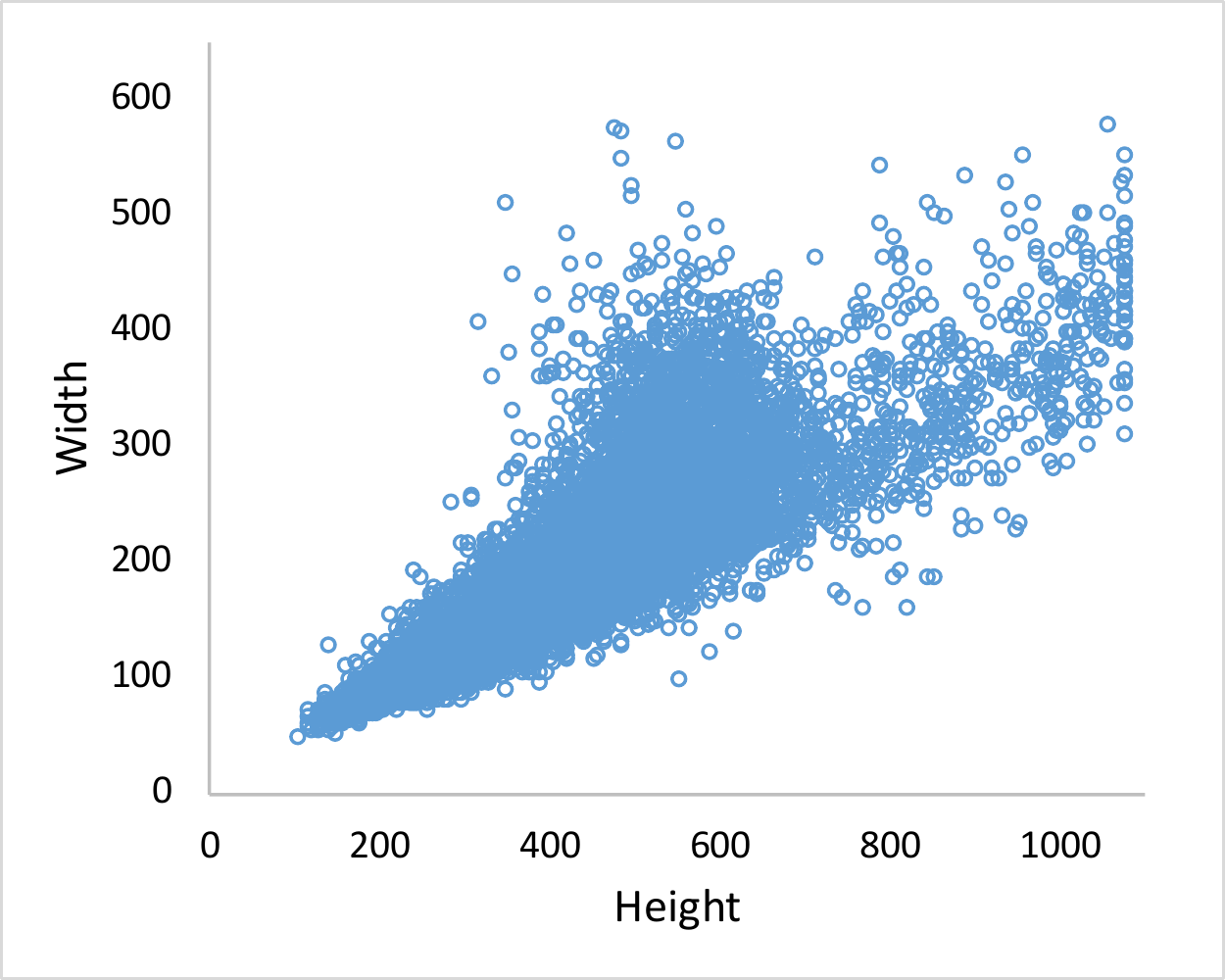}
  \centerline{(a) Statistics of frame sizes.}
  \end{minipage}
  \begin{minipage}[t]{0.3\linewidth}
  \centering
  \includegraphics[width=\linewidth]{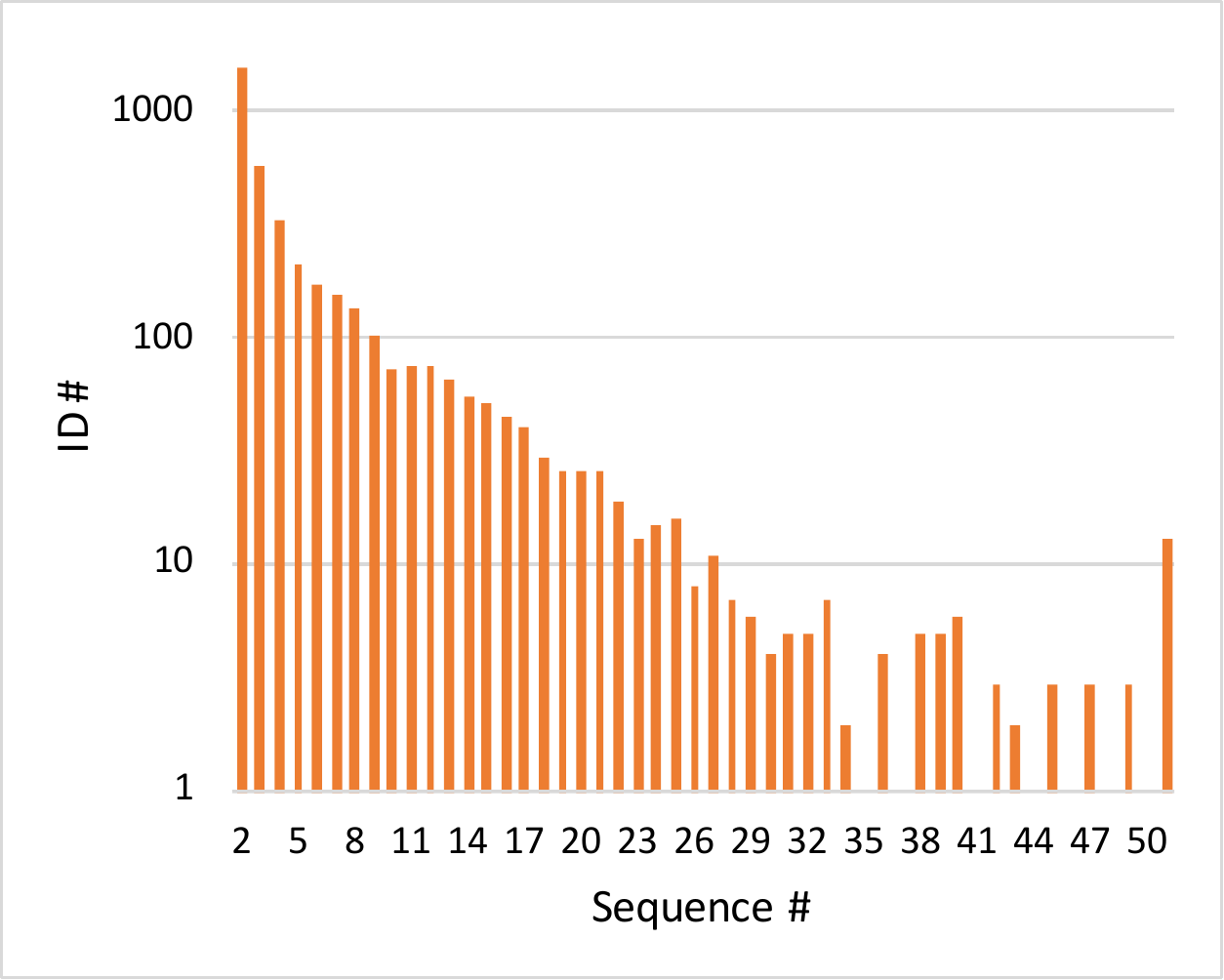}
  \centerline{(b) ID \# over sequence \#.}
  \end{minipage}
  \begin{minipage}[t]{0.3\linewidth}
  \centering
  \includegraphics[width=\linewidth]{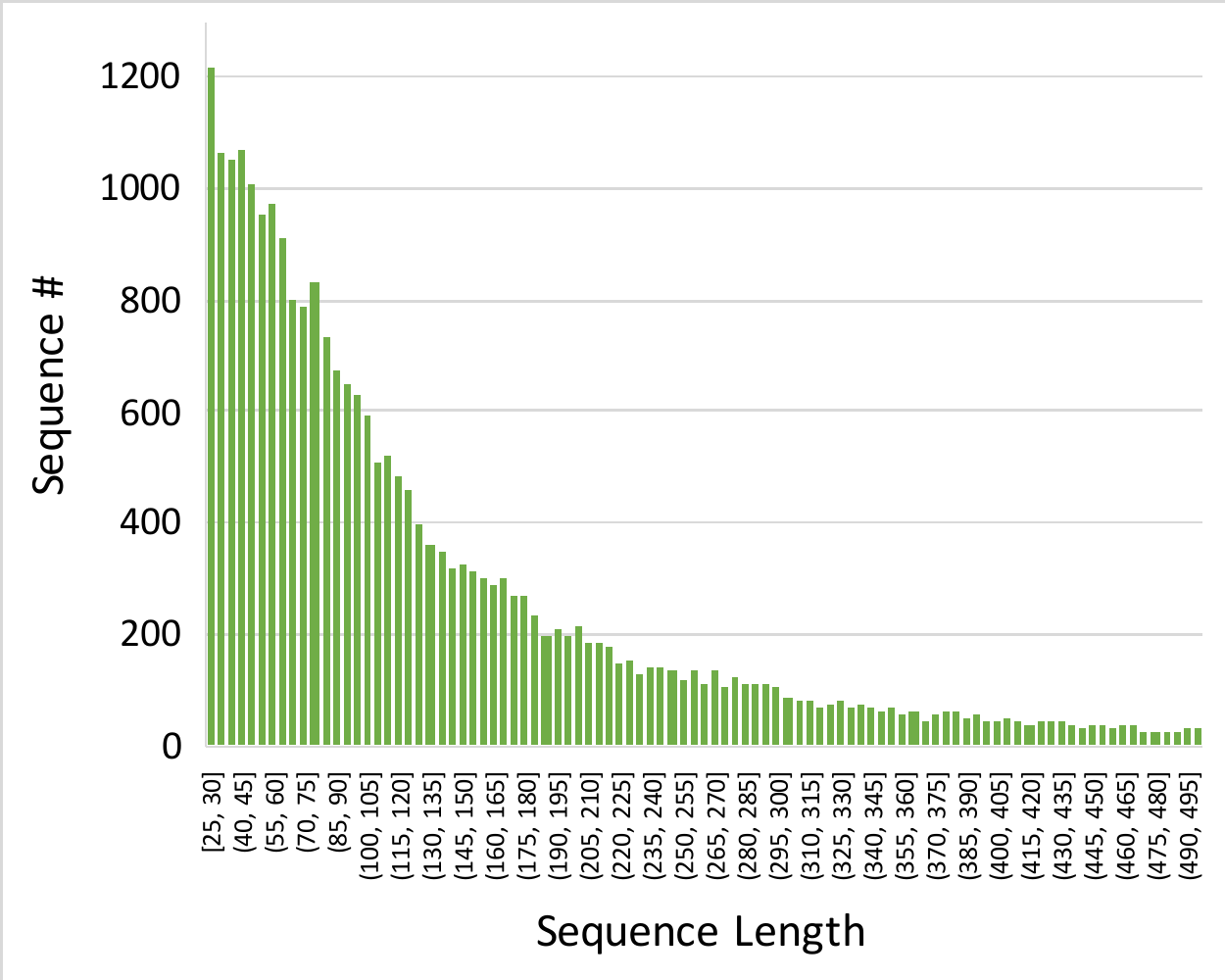}
  \centerline{(c) Sequence \# over sequence lengths.}
  \end{minipage}
   \caption{Statistics about the Gait3D dataset.}
   \label{fig:figure3} \vspace{-5mm}
\end{figure*}

\subsubsection{Generation of gait representations}
With the clean sequences of 4,000 IDs, we generate the 3D SMPL parameter, 3D mesh, 3D pose, 2D silhouette, and 2D pose for each frame.
For the 3D SMPL, 3D mesh, and 3D pose, we exploit a state-of-the-art 3D human mesh recovery method, ROMP~\footnote{\url{https://github.com/Arthur151/ROMP} under the MIT License.}~\cite{Sun_2021_ICCV}, since it can efficiently output these three representations in an end-to-end framework.
For the 2D silhouette, we use the semantic segmentation method, HRNet-segmentation~\footnote{\url{https://github.com/HRNet/HRNet-Semantic-Segmentation} under the MIT license.}~\cite{pami/00010CJDZ0MTW0X21}, to obtain the silhouette of the person in each frame.
For the 2D pose, we also utilize the HRNet to estimate the 2D keypoints of the person in each frame.
We keep the original resolution and aspect ratio of the frame without resizing or normalization.
Some examples of gait representations in our dataset are shown in Figure~\ref{fig:figure2}.
It is worth noting that we will only release the generated gait representations but not release any RGB frames to protect the privacy of the subjects.

\subsection{Dataset Statistics and Evaluation Protocol}
\label{subsec:statistics}
The statistics about the sizes of frames, ID numbers over sequence numbers, and sequence numbers over sequence lengths are shown in Figure~\ref{fig:figure3}.
From Figure~\ref{fig:figure3} (a), we can find that most frames range from $100 \sim 400 \times 200 \sim 800$ which are larger than person bounding boxes of existing datasets.
Figure~\ref{fig:figure3} (b) shows that most IDs have 2 $\sim$ 25 sequences, which guarantees the high reappearance times of subjects.
Figure~\ref{fig:figure3} (c) reflects that most sequences are longer than 50 frames (2 seconds) and the longest sequence has 500 frames, which reflects the complexity of the gait sequences in the unconstrained scenes.
The above statistics demonstrate that the Gait3D dataset is scalable but challenging for gait recognition research.

To facilitate the research, we split the 4,000 IDs of the Gait3D dataset into the train/test subsets with 3,000/1,000 IDs, respectively.
For the test set, we further randomly select one sequence from each ID to build the query set with 1,000 sequences, while the rest of the sequences become the gallery set with 5,369 sequences.
Our evaluation protocol is based on the open-set instance retrieval setting like existing gait recognition datasets~\cite{tifs/IwamaOMY12} and the person ReID task~\cite{iccv/ZhengSTWWT15}.
Given a query sequence, we measure its similarity between all sequences in the gallery set.
Then a ranking list of the gallery set is returned by the descending order of the similarities.
We report the average Rank-1 and Rank-5 identification rates over all query sequences.
We also adopt the mean Average Precision (mAP) and mean Inverse Negative Penalty (mINP)~\cite{corr/abs-2001-04193} which consider the recall of multiple instances and hard samples.

\vspace{-7mm}

\section{Experiments}
\label{sec:experiments}
In the experiments, we first evaluate several State-Of-The-Art (SOTA) 2D gait recognition methods and our SMPLGait on the Gait3D dataset.
Then, we analyze the influence of the frame size, the sequence length, and the scale of training IDs on the performance of gait recognition.

\begin{table*}[t]
\small
\centering
\begin{tabular}{l|l|cccc|cccc}
\multicolumn{2}{c|}{Input Size (W$\times$H)}  &\multicolumn{4}{c|}{88$\times$128}   & \multicolumn{4}{c}{44$\times$64} \\ 
Methods                      			& Publication & R-1 (\%) & R-5 (\%) & mAP (\%) & mINP & R-1 (\%) & R-5 (\%) & mAP (\%) & mINP \\ \midrule[1.5pt]
GEINet~\cite{icb/ShiragaMMEY16}      	& ICB 2016	& 7.00  & 16.30 & 6.05  & 3.77  & 5.40  & 14.20 & 5.06  & 3.14 \\
GaitSet~\cite{aaai/ChaoHZF19}		    & AAAI 2019	& 42.60 & 63.10 & 33.69 & 19.69	& 36.70 & 58.30 & 30.01 & 17.30	\\
GaitPart~\cite{cvpr/FanPC0HCHLH20}		& CVPR 2020	& 29.90 & 50.60 & 23.34 & 13.15	& 28.20 & 47.60 & 21.58 & 12.36	\\ 
GLN~\cite{eccv/HouCLH20}				& ECCV 2020	& 42.20 & 64.50 & 33.14 & 19.56	& 31.40 & 52.90 & 24.74 & 13.58	\\
GaitGL~\cite{Lin_2021_ICCV}	            & ICCV 2021	& 23.50 & 38.50 & 16.40 & 9.20	& 29.70 & 48.50 & 22.29 & 13.26	\\ 
CSTL~\cite{Huang_2021_ICCV}	            & ICCV 2021	& 12.20 & 21.70 & 6.44  & 3.28	& 11.70 & 19.20 & 5.59  & 2.59	\\ \midrule
PoseGait~\cite{pr/LiaoYAH20}	        & PR 2020	& 0.24  & 1.08  & 0.47  & 0.34	& - & - & -  & - \\ 
GaitGraph~\cite{corr/abs-2101-11228}	& arXiv 2021& 6.25  & 16.23 & 5.18  & 2.42	& - & - & -  & - \\  \midrule
SMPLGait  w/o 3D	            & Ours & 47.70 & 67.20 & 37.62 & 22.24 & 42.90 & 63.90 & 35.19 & 20.83	\\
SMPLGait       & Ours  & \textbf{53.20}	& \textbf{71.00} & \textbf{42.43} & \textbf{25.97} & \textbf{46.30} & \textbf{64.50} & \textbf{37.16} & \textbf{22.23}  \\  
\end{tabular} 
\caption{Comparison of the state-of-the-art gait recognition methods on Gait3D. As the inputs of the model-based methods, i.e., PoseGait and GaitGraph, are unrelated to the frame size, we only report one group of results.} \vspace{-5mm}
\label{tab:table2}
\end{table*}

\subsection{Evaluation of Existing Methods}
\label{subsec:evaluation}
Here, we evaluate eight SOTA 2D gait recognition methods including six model-free methods and two model-based methods.
We also compare our 3D gait recognition method (SMPLGait) with these methods.

\vspace{-3mm}
\subsubsection{Model-free Approaches}
\label{subsubsec:modelfreemethods}
The details of model-free approaches are as follows:

\textbf{1) GEINet~\cite{icb/ShiragaMMEY16}} is one of the first methods that adopts a four-layer CNN to learn gait features from GEIs using the cross-entropy loss.

\textbf{2) GaitSet~\cite{aaai/ChaoHZF19}} is the representative method that utilizes a 10-layer CNN to directly learn discriminative gait features from silhouette sequences. 
The GaitSet is trained by the batch all triplet loss~\cite{CoRR:BacthAllTriplet}.

\textbf{3) GaitPart~\cite{cvpr/FanPC0HCHLH20}} adopts the idea of multi-scale feature learning. 
It horizontally divides a silhouette image into fixed parts to learn discriminative micro-motion features.

\textbf{4) GLN~\cite{eccv/HouCLH20}} is an efficient and effective method to learn compact features from gait sequences, which achieves the SOTA performance using only a 256-D feature.

\textbf{5) GaitGL~\cite{Lin_2021_ICCV}} is also a CNN-based framework to learn both global and local features from gait sequences.

\textbf{6) CSTL~\cite{Huang_2021_ICCV}} applies multi-scale learning on the temporal dimension of the sequence to learn both long-term and short-term motion for gait recognition.

\textbf{Implementation Details:}
During training, we train the above models except GLN with the same configuration.
The batch size is $32\times4\times30$, where 32 denotes the number of IDs, 4 denotes the number of training samples per ID, and 30 is the sequence length.
The models are trained for 1,200 epochs with the initial Learning Rate (LR)=1e-3 and the LR is multiplied by 0.1 at the 200-th and 600-th epochs.
The optimizer is Adam~\cite{corr/KingmaB14} and the weight decay is set to 5e-4.
For GLN, we follow the two-stage training as in~\cite{eccv/HouCLH20}.
The model trained in the first stage is used as the pre-trained model for the second stage.
Both of the two stages are trained with the same configuration of other methods.
During testing, we use the cosine similarity to measure the similarity between each pair of query and gallery sequences.
For the GaitSet, GaitPart, GLN, and GaitGL models, we adopt the implementations in the open-source OpenGait toolbox~\footnote{\url{https://github.com/ShiqiYu/OpenGait}} since they outperform the original codes.

\vspace{-3mm}
\subsubsection{Model-based Approaches}
\label{subsubsec:modelbasedmethods}
\vspace{-1mm}
We compare two representative model-based methods which use 2D or 3D skeletons as the input.

\textbf{1) PoseGait~\cite{pr/LiaoYAH20}} first exploits OpenPose~\cite{pami/CaoHSWS21} to extract the 2D keypoints from RGB frames, then uses the method in~\cite{cvpr/ChenR17} to estimate the 3D keypoints of human bodies.
Based on the 3D skeletons, it defines several parameters such as joint angle, limb length, and joint motion together with the pose features as the gait representation.
In our implementation, we train it for 700 epochs with a batch size of 128. 
The LR is set to 1e-3. 
The optimizer is Adam~\cite{corr/KingmaB14} and weight decay is equal to 5e-4.

\textbf{2) GaitGraph~\cite{corr/abs-2101-11228}} is a recent model-based gait recognition method.
It models the 2D skeleton as a graph and adopts a Graph Convolution Network, i.e., the ResGCN~\cite{mm/Song0SW20}, to learn features by the contrastive loss.
We train GaitGraph in two stages.
The setting of the first stage is the same as PoseGait, and the model trained in the first stage is used as the pre-trained model of the second stage. 
In the second stage, we fine-tune it for 250 epochs.

\vspace{-3mm}
\subsubsection{Implementation Details of the SMPLGait}
\label{subsubsec:ourmethod}
\vspace{-1mm}
For our SMPLGait, we use the loss in Equ.~\ref{equ4} for training.
In 3D-STN, we set the dropout rate to 0.2 for FC layers.
The hyper-parameters in Equ.~\ref{equ4} are set as $\alpha$=1.0 and $\beta$=0.1.
Other settings are the same as those in Section~\ref{subsubsec:modelfreemethods}.

\vspace{-3mm}
\subsubsection{Experimental Results}
\label{subsubsec:sotamethodsanalysis}
The results of model-free methods, model-based methods, and our SMPLGait are listed in Table~\ref{tab:table2}.

For model-free methods, we can first observe that the overall performance of the SOTA methods is much worse than their performance on in-the-lab datasets like the CASIA-B~\cite{icpr/YuTT06} and OU-ISIR series~\cite{tifs/IwamaOMY12, ipsjtcva/TakemuraMMEY18}.
This reflects that there is a huge gap between the in-the-lab research and the in-the-wild application that is much more challenging.
Meanwhile, the performance of the SOTA model-free methods varies significantly.
For example, the GEI-based method, i.e., GEINet obtains the worst results, which indicates that the GEIs discard too much useful information for gait recognition.
Moreover, the methods considering the order of the frames in sequences, i.e., GaitPart, GLN, GaitGL, and CSTL, obtain lower accuracy. 
It means that the temporal information in the wild scene is hard to learn, because people may stop then continue to walk at varying speeds and routes in unconstrained scenarios.
On the contrary, the methods considering frames as an unordered set, i.e., GaitSet, obtain better results.

For model-based methods, we can find that they are greatly worse than model-free methods on the Gait3D dataset. 
This is because the input of the model-based methods only has a few sparse human body joints, which seriously lacks useful gait information, such as body shape, appearance, and so on.
In addition, the walking speed and route are uncertain in real scenarios, which also greatly affects the performance of the model-based methods that aim to model the temporal dynamics of the human body.

Finally, our SMPLGait outperforms other methods by a large margin, which indicates the potential of 3D representations for gait recognition in the wild.

\subsubsection{Ablation Study of SMPLGait}
We also conduct an ablation study on the key components in SMPLGait by removing the 3D branch (SMPLGait w/o 3D).~\footnote{It should be noticed that SMPLGait w/o 3D is equal to OpenGait Baseline~\cite{opengait21}}
The results are listed in Table~\ref{tab:table2}.
This comparison shows that the integration of 2D and 3D representations can better address the challenges of gait recognition in the wild.

\subsection{More Analysis of the Gait3D Dataset}
\label{subsec:analysis}
We choose two SOTA gait recognition methods, i.e., GaitPart and GaitSet, and our SMPLGait (Ours) to analyze the influence of input size, frame number in sequences, and training ID number on the accuracy.
All the models are evaluated on the whole Gait3D test set.

\textbf{Input Size.}
We explore two input sizes of $88 \times 128$ and $44\times 64$ for the compared methods, as shown in Table~\ref{tab:table2}.
From the results, we can observe that the performance of almost all methods is improved with larger input size.
There is an exception, i.e., GaitGL, which obtain worse accuracy with larger input size.
This may be because that GaitGL adopts the 3D CNN as the backbone.
When using a larger input size, the 3D CNN learns more misalignment information about frames in physical space, which makes it more difficult to be optimized. 

\begin{figure}[t]
  \centering
   \includegraphics[width=0.9\linewidth]{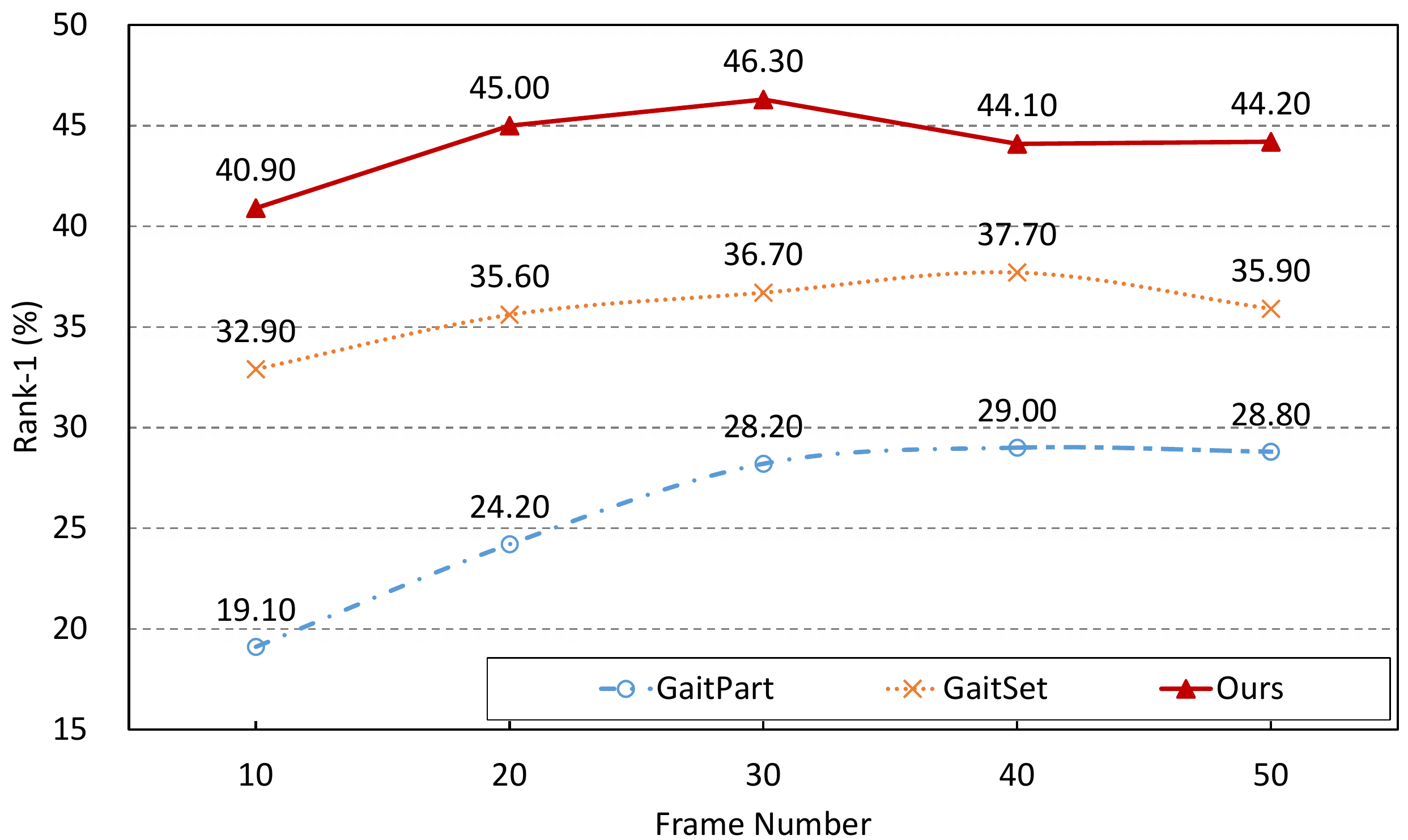}
   \caption{The effect of frame numbers in sequences.}
   \label{fig:figure5} \vspace{-5mm}
\end{figure}

\textbf{Number of Training Frames}
We randomly sample 10$\sim50$ frames from original gait sequences during training.
The Rank-1 accuracy is illustrated in Figure~\ref{fig:figure5}.
The results show that as the number of frames increases the performance first increases and then decreases, while the best performance occurs around 30 frames per sequence.
This indicates that more frames could not bring higher accuracy.
The reason may be that there is a lot of redundant or noisy information caused by uncertain speeds and routes of persons, which will bring ambiguous features for gait recognition.

\begin{figure}[t]
  \centering
   \includegraphics[width=0.9\linewidth]{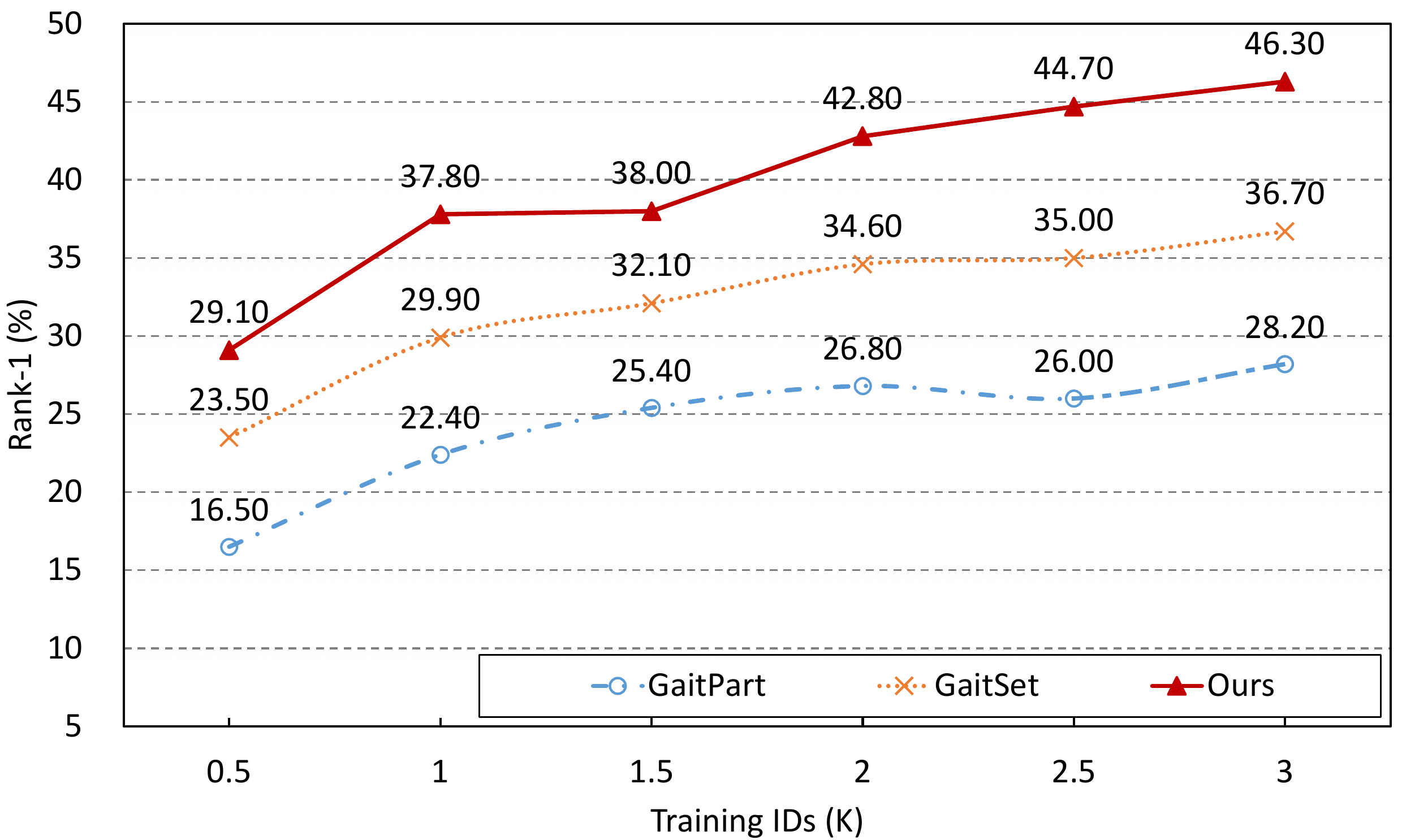}
   \caption{The effect of different training ID numbers.}
   \label{fig:figure6} \vspace{-5mm}
\end{figure}

\textbf{Scale of Training IDs}
We fix other settings and use 0.5K $\sim$ 3KIDs with an increment of 0.5K for training.
As shown in Figure~\ref{fig:figure6}, the performance of the models grows stably with more training IDs.
These results reflect the scalability of our Gait3D dataset.

More experiments and exemplar results on Gait3D can be found in \textbf{the supplementary material}.

\vspace{-1mm}
\section{Discussion}
\label{sec:discussion}
\vspace{-1mm}

\textbf{Ethical Issues.}
There are two main ethical issues of this paper: 1) privacy, and 2) data bias.
For the first issue, we will try our best to protect the privacy of the subjects involved in our dataset.
Firstly, we will not release any human cognizable data like original videos, RGB frames, and bounding boxes of persons.
Second, the dataset will be distributed only for research purposes via the case-by-case application with a strict license.
To eliminate data bias, the genders and ages of subjects are relatively balanced.

\textbf{Future Work.}
Despite the proposed baseline method for 3D gait recognition, there are many potential directions for this challenging task.
For example, one direction is to study how to design a deep CNN for learning more discriminative features directly from 3D meshes.
The second direction is how to learn the temporal information of gait representation, because the walking speed and route in Gait3D are irregular, it is significantly different from the datasets built in the lab. 
Another interesting direction is how to fuse the multi-modal information like silhouette, 2D/3D skeleton, and 3D mesh for gait recognition in the wild.

More discussions about the limitations and potential negative impact can be found in \textbf{the supplementary material}.

\vspace{-5mm}
\section{Conclusion}
\label{sec:conclusion}

Gait recognition in the wild faces significant challenges such as extreme viewpoint changes, occlusions of the human body, and complex clutter in the environment.
Existing methods using 2D silhouettes or skeletons will fail in the wild because crucial information like 3D viewpoints and shapes of human bodies is discarded.
Therefore, this paper proposes a 3D SMPL model-based framework (SMPLGait) which is the first method to explore dense 3D representations for gait recognition in the wild.
To facilitate the research, we build the first large-scale 3D gait recognition dataset (Gait3D) from cameras deployed in a large supermarket.
It provides diverse gait representations including 3D meshes, 3D SMPLs, 3D poses, 2D silhouettes, and 2D poses for over 25,000 gait sequences of 4,000 subjects.
We hope Gait3D can provide researchers with a new perspective on gait recognition.

\textbf{Acknowledgements.} This work was supported in part by the National Key Research and Development Program of China under Grant 2020AAA0103800, in part by the National Nature Science Foundation of China under Grant 61931008 and Grant U21B2024.

\appendix

\section{Appendix: Details of the SMPLGait}
\label{sec:detials}

\subsection{Details of the Network Structure}
\label{subsec:network}

\textbf{The Silhouette Learning Network} (SLN) consists of six convolutional layers, and each convolutional layer is followed by LeakyReLU whose negative slope is equal to 0.01. 
The structure of SLN is inspired by GaitSet~\cite{aaai/ChaoHZF19} and OpenGait~\footnote{\url{https://github.com/ShiqiYu/OpenGait}}. 
The detailed parameters are listed in Table~\ref{tab:table_sln_structure}.

\begin{table}[h]
\footnotesize
\centering
\begin{tabular}{l|ccccc}
Layers              & Kernel \#   & Kernel Size       & Stride    & Padding    \\ \midrule[1.5pt]
Conv1	            & 64        & 5$\times$5        & 1         & 2       & \\ 
LeakyReLU (0.01)	& -         & -                 & -         & -       & \\ \midrule
Conv2	            & 64        & 3$\times$3        & 1         & 1       & \\ 
LeakyReLU (0.01)    & -         & -                 & -         & -       & \\
Max Pooling         & -         & 2$\times$2        & 2         & 0       & \\ \midrule
Conv3               & 128       & 3$\times$3        & 1         & 1       & \\ 
LeakyReLU (0.01)    & -         & -                 & -         & -       & \\ \midrule
Conv4   	        & 128       & 3$\times$3        & 1         & 1       & \\ 
LeakyReLU (0.01)    & -         & -                 & -         & -       & \\
Max Pooling	        & -         & 2$\times$2        & 2         & 0       & \\ \midrule
Conv5   	        & 256       & 3$\times$3        & 1         & 1       & \\ 
LeakyReLU (0.01)    & -         & -                 & -         & -       & \\ \midrule
Conv6   	        & 256       & 3$\times$3        & 1         & 1       & \\
LeakyReLU (0.01)    & -         & -                 & -         & -       & \\
\end{tabular}
\caption{Details of the SLN Network.}
\label{tab:table_sln_structure}
\end{table}

\textbf{The 3D Spatial Transformation Network} (3D-STN) consists of several Fully Connected (FC) layers with dropout, Batch Normalization (BN) layers, and ReLU activation functions. 
The details of 3D-STN are listed in Table~\ref{tab:table_stn_structure}. 
Note that the $h$ and $w$ listed in the table are the height and width of the feature map from SLN.
For convenient computation, we set $h=w=max(h, w)$ in our experiments.

\begin{table}[h]
\footnotesize
\centering
\begin{tabular}{l|cc}
Layers              & Neuron \#     & Dropout Rate  \\ \midrule[1.5pt]
FC1	                & 128           & 0.0           \\ 
BN1                 & -             & -             \\ 
ReLU            	& -             & -             \\ \midrule
FC2	                & 256           & 0.2           \\ 
BN2                 & -             & -             \\ 
ReLU            	& -             & -             \\ \midrule
FC3	                & $h \times w$  & 0.2           \\ 
BN3                 & -             & -             \\ 
ReLU            	& -             & -             \\
\end{tabular}
\caption{Details of the 3D-STN Network. (Input Size: 88$\times$128).}
\label{tab:table_stn_structure}
\end{table}

\section{Appendix: Additional Experimental Results}
\label{sec:additional}
In this section, we first analyze the effect of sequence length on accuracy during inference.
Then, we conduct the cross-domain experiments to reflect the domain gap between existing datasets and our Gait3D dataset.
At last, we provide some exemplar results of our SMPLGait framework for gait recognition to qualitatively demonstrate the effectiveness of our method.

\subsection{Effect of the Lengths of Test Sequences}
\begin{figure}[t]
  \centering
  \includegraphics[width=0.99\linewidth]{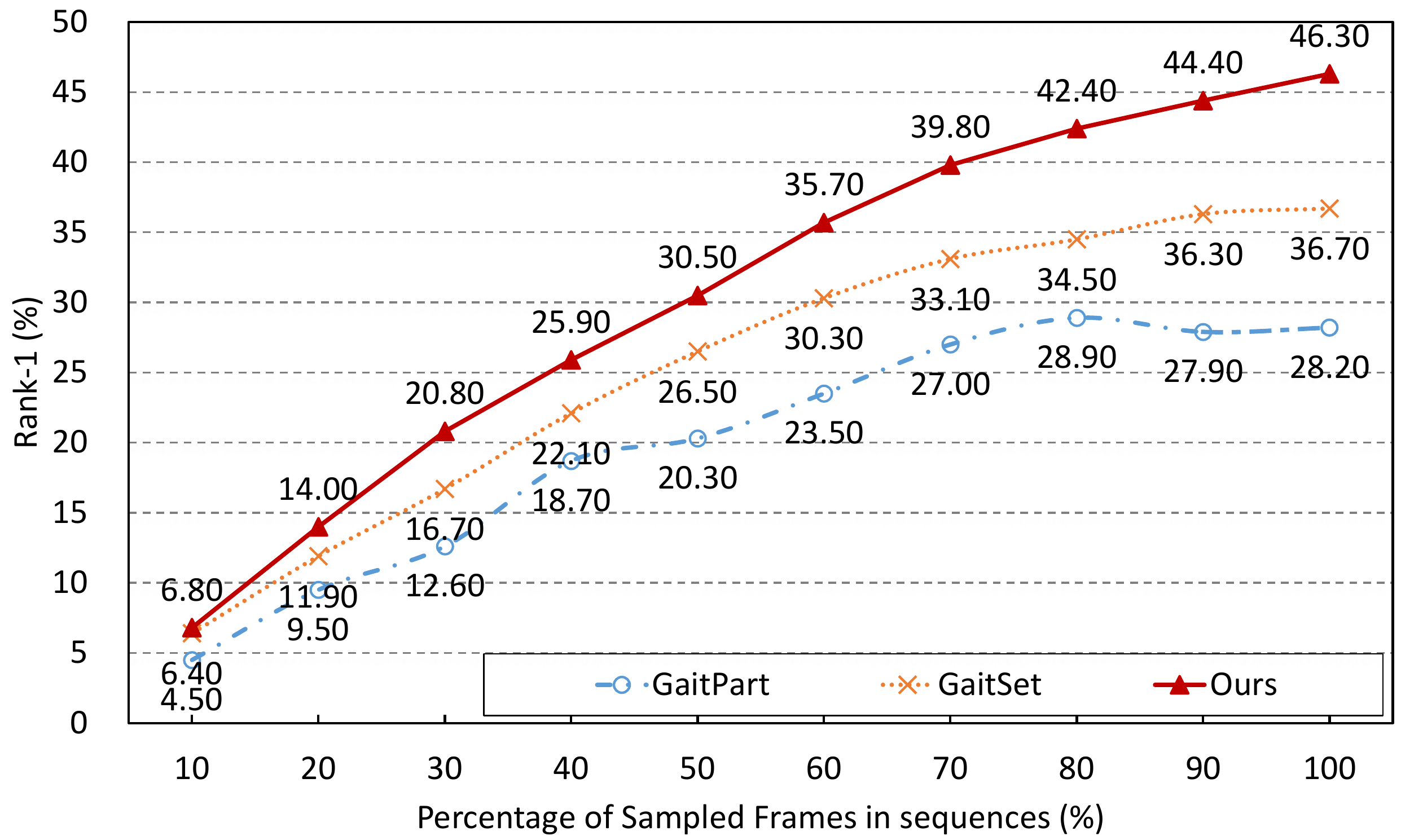}
  \caption{The effect of the lengths of test sequences.}
  \label{fig:figure_test_seq_length}
\end{figure}

This subsection analyzes the influence of the length of testing sequences on the accuracy of gait recognition.
We sample 10\% $\sim$ 100\% frames with the 10\% increment from the sequences during testing.
The plots are demonstrated in Figure~\ref{fig:figure_test_seq_length}.
From the results, we can observe that the accuracy is improved with the increasing frames of the sequence.
Therefore, in real practice, we need to make a trade-off between accuracy and efficiency.

\subsection{Cross-domain Experiments}

In this subsection, we analyze the domain gap between our Gait3D dataset and existing widely used datasets including CASIA-B~\cite{icpr/YuTT06}, OU-LP~\cite{tifs/IwamaOMY12} and the recently released GREW~\cite{Zhu_2021_ICCV}.
Because existing datasets do not provide 3D representations, we only adopt the 2D silhouettes for gait representations in our experiments.
We adopt the GaitSet~\cite{aaai/ChaoHZF19} for the cross-domain experiments since it is the SOTA model-free method on the Gait3D dataset.

\subsubsection{Evaluation Protocol}

In the cross-domain experiments, we train the GaitSet model on the training set of one source dataset and evaluation the trained model on the testing set of one target dataset.
For CASIA-B, OU-LP, and Gait3D, we use the official train/test split for training and testing.
Because the test set of GREW~\cite{Zhu_2021_ICCV} is not released publicly for evaluation, we randomly sample 1,000 IDs from the training set of GREW for evaluation.
For the sampled 1,000 IDs, we further randomly select one sequence from each ID to build the query set with 1,000 sequences, while the rest of the sequences becomes the gallery set with 4,095 sequences.
We utilize Rank-1 (R-1), Rank-5 (R-5), and mAP as the evaluation metrics.

\subsubsection{Main Results}

The results of the cross-domain evaluation are listed in Table~\ref{tab:table_cross_domain}.
From the results, we first find that the GaitSet models trained on the in-the-lab datasets, i.e., CASIA-B~\cite{icpr/YuTT06} and OU-LP~\cite{tifs/IwamaOMY12} obtain very poor results, only 6.90\% and 6.10\% Rank-1.
This reflects that there is a huge domain gap between the in-the-lab research and the in-the-wild application.

Next, we observe that the model trained on GREW then tested on Gait3D only obtains 16.50\% Rank-1, while the model trained on Gait3D then tested on GREW achieves a much higher Rank-1, i.e., 43.86\%.
It is worth noting that the training set of GREW has 20,000 IDs, while our Gait3D only contains 3,000 IDs.
This demonstrates the significant domain gap between our Gait3D and GREW although the two datasets are both collected in the wild.
Moreover, it indicates that the model trained on Gait3D has a more powerful capability of generalization than the one trained on GREW.

At last, we can see that the GaitSet trained on Gait3D achieves competitive accuracy on in-the-lab datasets, i.e.,  Rank-1 of 66.71\% on CASIA-B and 97.84\% on OU-LP which is close to the model trained on OU-LP (99.89\%) in our implementation).
These results further prove that there is a huge gap between the in-the-lab dataset and in-the-wild application, while our Gait3D enables the model to learn more generalized gait representations.

\begin{table}[t]
\footnotesize
\centering
\begin{tabular}{l|l|cccc}
Source                          			& Target                    				& R-1 (\%) & R-5 (\%) & mAP (\%) \\ \midrule[1.5pt]
CASIA-B~\cite{icpr/YuTT06}	    	& \multirow{3}{*}{Gait3D}    		& 6.90 & 14.60 & 4.64 & \\ 
OU-LP~\cite{tifs/IwamaOMY12}	&                           				& 6.10 & 12.40 & 4.42 & \\
GREW~\cite{Zhu_2021_ICCV}	    	&                           				& 16.50 & 31.10 & 11.71 & \\ \midrule
\multirow{3}{*}{Gait3D}  			& CASIA-B~\cite{icpr/YuTT06}	    	& 66.71 & 71.59 & 33.88 & \\ 
                        					& OU-LP~\cite{tifs/IwamaOMY12}	& 97.84 & 99.38 & 68.06 & \\
                        					& GREW~\cite{Zhu_2021_ICCV}	& 43.86 & 60.89 & 28.06 & \\
\end{tabular}
\caption{Results of cross-domain experiments. The method is trained on each source dataset and directly tested on the target datasets.}
\label{tab:table_cross_domain}
\end{table}

\subsection{Exemplar Results of SMPLGait}
\label{subsubsec:sotamethodsanalysis}
\begin{figure*}[t]
  \centering
  \includegraphics[width=1.0\linewidth]{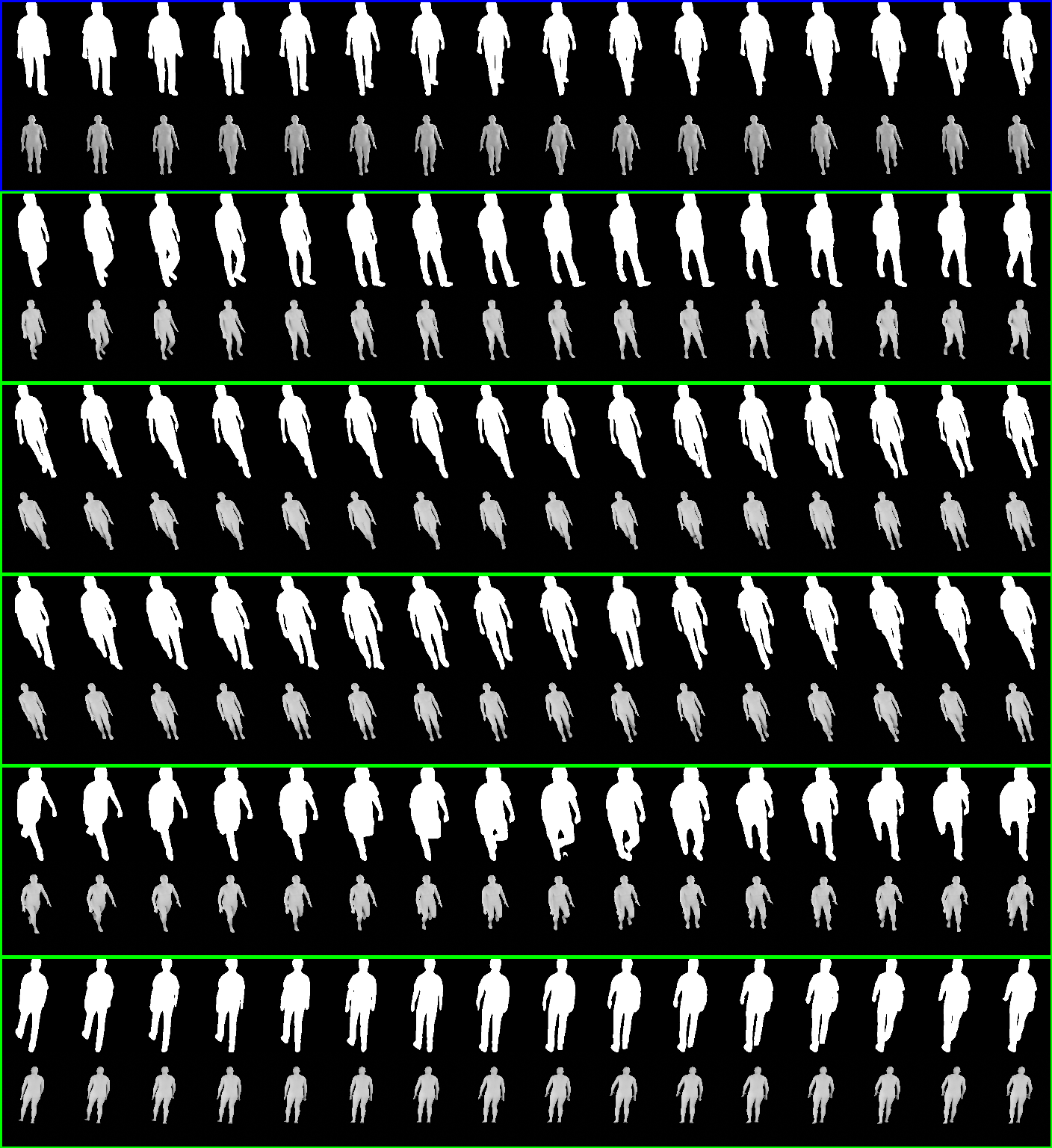}
    \caption{Exemplar results of SMPLGait on the Gait3D. 16 consecutive frames are sampled from each sequence for visualization. This case shows that our method obtains good results when the samples are high-quality. (Best viewed in color.)} \vspace{-3mm}
  \label{fig:figure_vis_1}
\end{figure*}

\begin{figure*}[t]
  \centering
  \includegraphics[width=1.0\linewidth]{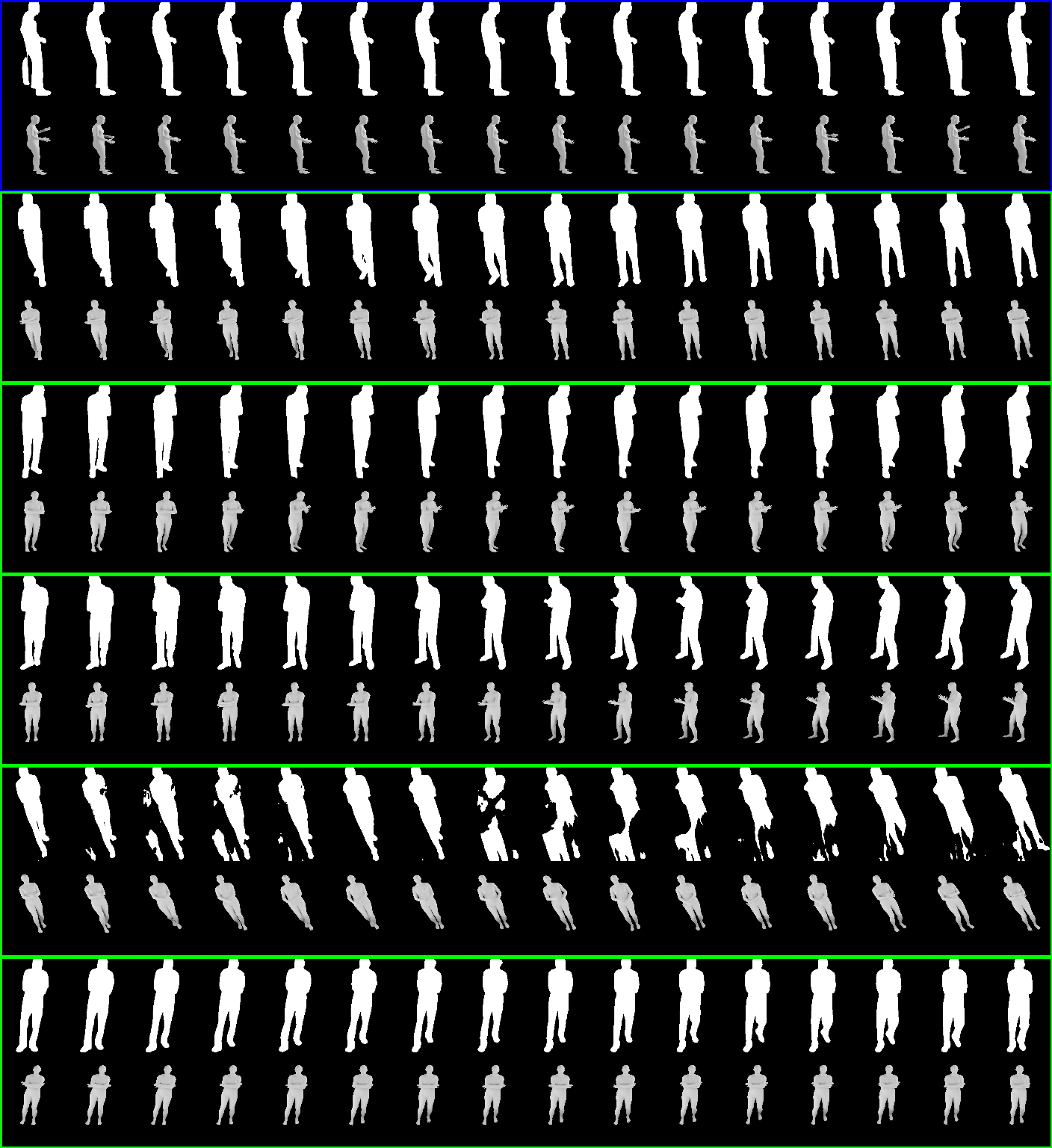}
  \caption{Exemplar results of SMPLGait on the Gait3D. This example reflects that our method can work well when part of the person is occluded. (Best viewed in color.)}
  \label{fig:figure_vis_2}
\end{figure*}

\begin{figure*}[t]
  \centering
  \includegraphics[width=1.0\linewidth]{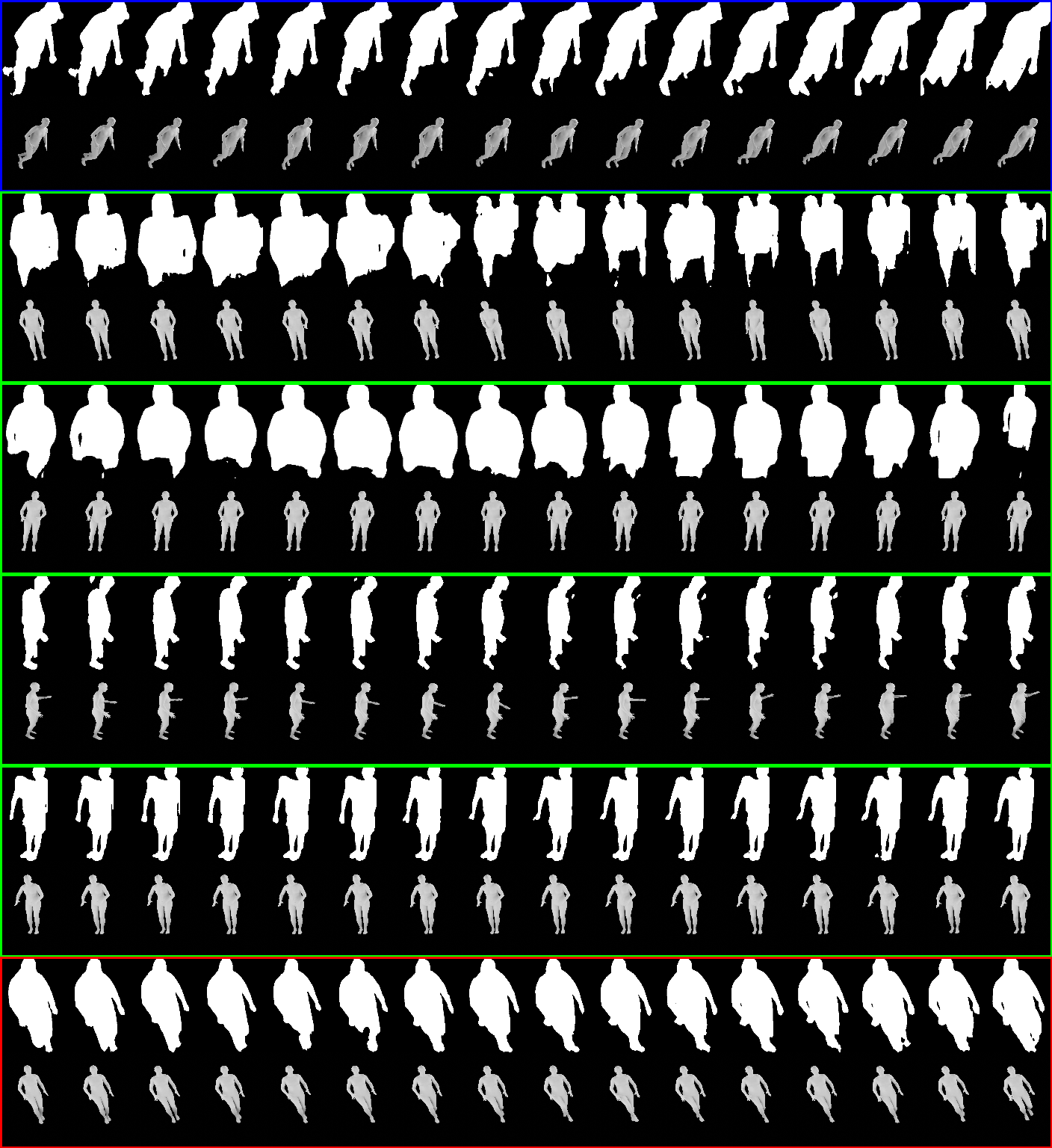}
    \caption{Exemplar results of SMPLGait on the Gait3D. This example shows that even when the silhouettes are of low quality, the 3D meshes can help the model obtain correct results. (Best viewed in color.)}
  \label{fig:figure_vis_3}
\end{figure*}

\begin{figure*}[t]
  \centering
  \includegraphics[width=1.0\linewidth]{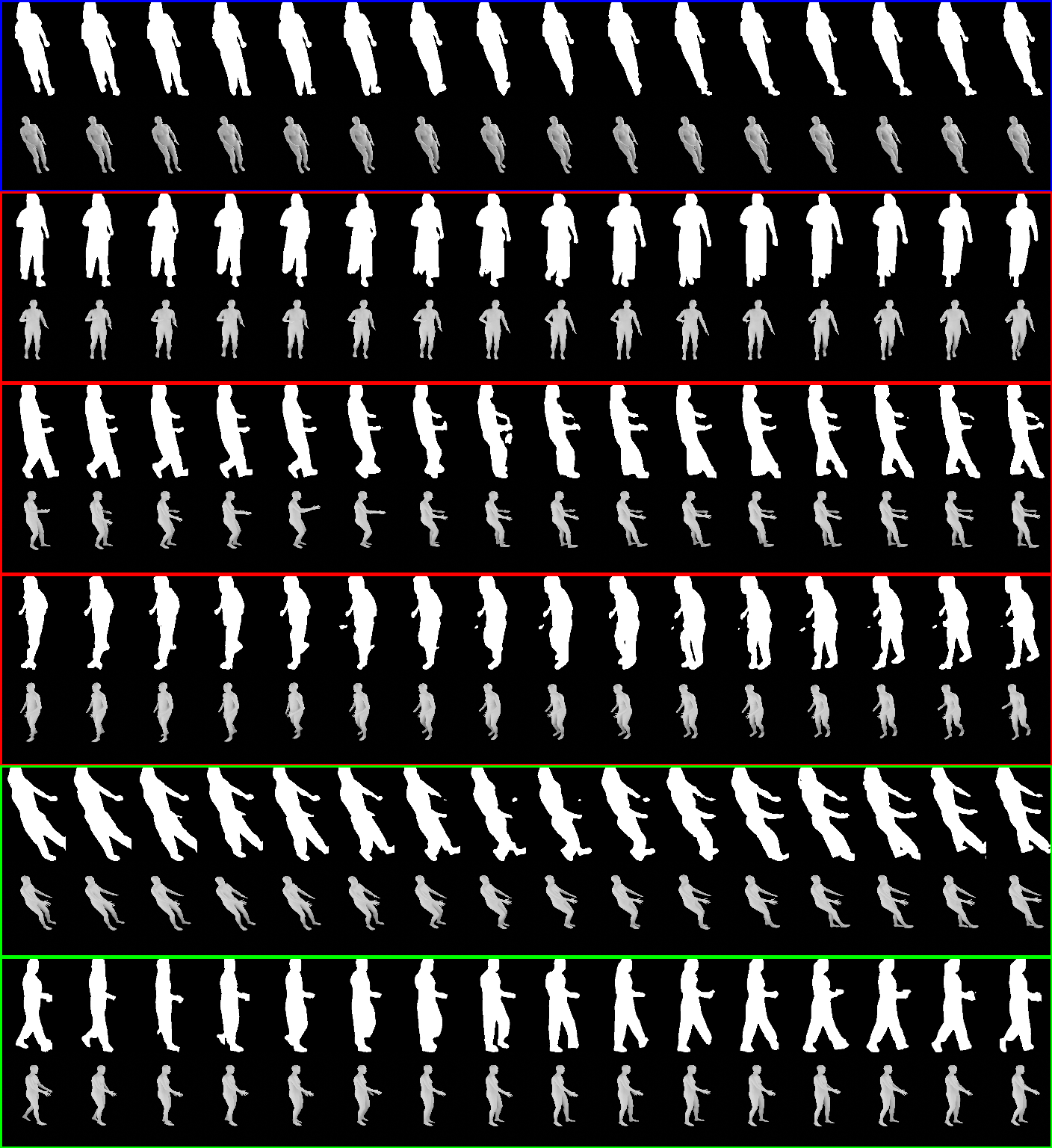}
    \caption{A bad case of SMPLGait on the Gait3D. The top-3 results contain persons wearing very similar clothes to the query, which is a very challenging condition of gait recognition. (Best viewed in color.)}
  \label{fig:figure_vis_4}
\end{figure*}

Figure~\ref{fig:figure_vis_1} -~\ref{fig:figure_vis_4} provides several exemplar results of our SMPLGait framework on the Gait3D dataset.
The top two rows with \textcolor{blue}{\textbf{blue}} bounding boxes are the silhouette sequence and 3D Mesh sequence of the query, respectively.
The rows following the query are the top-5 gallery sequences ranked by their similarities to the query sequence.
The results with \textcolor{green}{\textbf{green}} bounding boxes are the correctly matched sequences, while those with \textcolor{red}{\textbf{red}} bounding boxes are wrong results.

From Figure~\ref{fig:figure_vis_1} -~\ref{fig:figure_vis_3}, we can observe that the 3D representations can work well in multi-viewpoint, occluded, and multi-person cases.
By this means, 3D meshes can provide more information about shapes, poses, and viewpoints of human bodies, which can help improve the accuracy of gait recognition in real-world scenes.

Figure~\ref{fig:figure_vis_4} illustrates a bad case in which the top-3 persons with similar clothes and shapes seriously interfere with the matching results.
This indicates that very similar clothing and shapes of persons are one of the main challenges of gait recognition.


\section{Appendix: Discussion}
\label{sec:discussion}

\textbf{Limitations.}
Although we proposed a 3D gait recognition framework, its performance still has a large space to improve for practical applications.
In addition, we only exploit a few frames of gait sequences, e.g., 30 frames, in our framework.
The temporal dynamics are not fully explored for gait recognition in the wild.

\textbf{Potential Negative Impact.} The potential negative outcomes mainly come from the fact that, with the large-scale deployment of the urban monitoring network, if the accuracy of gait recognition in the real scenarios is greatly improved in the future, it may cause some privacy and security issues.
To minimize these risks, our Gait3D dataset will be distributed only for research purposes via the case-by-case application with a strict license.
In summary, we will try our best to protect the privacy of the subjects.
We hope that the development of gait recognition can help to create a better human society, as well as better services for human beings, such as assisting the police to solve crimes, looking for lost people, and so on.

{
\small
\balance
\bibliographystyle{ieee_fullname}
\bibliography{cvpr2022}
}

\end{document}